\documentclass[final]{cvpr}

\usepackage{times}
\usepackage{epsfig}
\usepackage{graphicx}
\usepackage{amsmath}
\usepackage{amssymb}
\usepackage{algorithm}
\usepackage{algpseudocode}
\usepackage{multirow}
\usepackage{pifont}
\usepackage{booktabs}
\usepackage{epsfig}
\usepackage{booktabs,multirow}
\usepackage{graphics}
\usepackage{threeparttable}
\usepackage{color}
\usepackage[normalem]{ulem}
\usepackage{multirow}
\usepackage{float}
\usepackage{amsfonts}
\usepackage{bm}
\usepackage{rotating}
\usepackage{enumitem}
\usepackage[numbers]{natbib}
\usepackage{array}
\usepackage[table]{xcolor}
\usepackage{colortbl}
\definecolor{myy}{RGB}{126,95,0}
\definecolor{mygray}{gray}{.9}
\definecolor{bblue}{RGB}{30,80,120}
\definecolor{mygray1}{gray}{.7}
\usepackage{bm}
\usepackage{mathtools}
\usepackage{subcaption}
\usepackage{siunitx}
\usepackage[nopar]{lipsum}
\usepackage[export]{adjustbox}

\newcolumntype{I}{!{\vrule width 1pt}}

\definecolor{ggray}{RGB}{127,127,127}
\definecolor{mygreen}{RGB}{93,174,86}
\definecolor{myred}{RGB}{192,0,0}

\makeatletter
\newcommand{\thickhline}{%
	\noalign {\ifnum 0=`}\fi \hrule height 1pt
	\futurelet \reserved@a \@xhline
}
\makeatother
\newcommand{\tabincell}[2]{\begin{tabular}{@{}#1@{}}#2\end{tabular}}

\usepackage{caption}
\usepackage[pagebackref=true,breaklinks=true,colorlinks,bookmarks=false]{hyperref}
\usepackage[utf8]{inputenc}

\usepackage{cleveref}
\crefname{section}{§}{§§}
\Crefname{section}{§}{§§}

\usepackage{lipsum}

\newcommand{\myparagraph}[1]{\vspace{1pt}\noindent{\bf #1}}

\begin{document}


\title{MTR-A: 1st Place Solution for \\2022 Waymo Open Dataset Challenge - Motion Prediction}

\author{
Shaoshuai Shi ~~~~Li Jiang ~~~~Dengxin Dai ~~~~Bernt Schiele 
\vspace{0.2cm}\\
Max Planck Institute for Informatics, Saarland Informatics Campus\\
{\tt\small \{sshi, lijiang, ddai, schiele\}@mpi-inf.mpg.de}
}

\maketitle

\begin{abstract}
In this report, we present the 1st place solution for motion prediction track in 2022 Waymo Open Dataset Challenges. 
We propose a novel Motion Transformer framework for multimodal motion prediction, which introduces a small set of novel motion query pairs for generating better multimodal future trajectories by jointly performing the intention localization and iterative motion refinement.   
A simple model ensemble strategy with non-maximum-suppression is adopted to further boost the final performance. 
Our approach achieves the 1st place on the motion prediction leaderboard of 2022 Waymo Open Dataset Challenges, outperforming other methods with remarkable margins. Code will be available at \url{https://github.com/sshaoshuai/MTR}.

\end{abstract}
\section{Introduction} \label{sec:intro}
Recently, motion prediction is receiving increasing attention~\cite{gao2020vectornet, liang2020learning, gu2021densetnt,tolstaya2021identifying,liu2021multimodal,ngiam2021scene,jia2022hdgt} as it is crucial for autonomous vehicles to make safe decisions.
It is also a highly challenging task due to its inherently multimodal behaviors of the agent and complex scene environments. 

To predict accurate future trajectories of the agent, existing approaches mainly follow two different lines.  Some approaches~\cite{zhao2020tnt,gu2021densetnt} adopt the goal-based strategy to localize the agent's destination with densely sampled goal candidates, which alleviate the burden of model optimization by reducing the trajectory uncertainty. 
Some other approaches~\cite{ngiam2021scene,varadarajan2021multipath++} direct predict a set of future trajectories based on the encoded agent feature, which can adaptively cover the agent's future behavior in a more flexible manner. 
However, the goal-based methods suffer from high computation and memory cost since their performance depends on a large number of goal candidates, while the direct-regression methods generally converge slowly as various motion modes are required to be regressed from the same agent features without any spatial priors. 
Hence, to address these limitations, we propose a novel framework, namely Motion Transformer (MTR), which takes the best of both worlds.

Specifically, our approach adopt a transformer encoder-decoder structure for multimodal motion prediction, where a small set of novel motion query pairs is proposed to model the multimodal future behaviors of the agent.
Each motion query pair contains a static intention query and a dynamic searching query, where the static intention query takes charge of predicting the future trajectory for a specific motion mode based on its associated spatial intention point, and the dynamic searching query conducts iterative motion refinement by continually aggregating trajectory-specific features.   
Thanks to these learnable and mode-specific motion query pairs, our framework not only stabilizes the training process by introducing spatial priors based on a small set of intention points, but also enables an adaptive prediction of future trajectory for each motion mode by retrieving their trajectory-specific feature.

\begin{figure*}[t]
	\centering
	\vspace{-3mm}
	\includegraphics[width=0.98\linewidth]{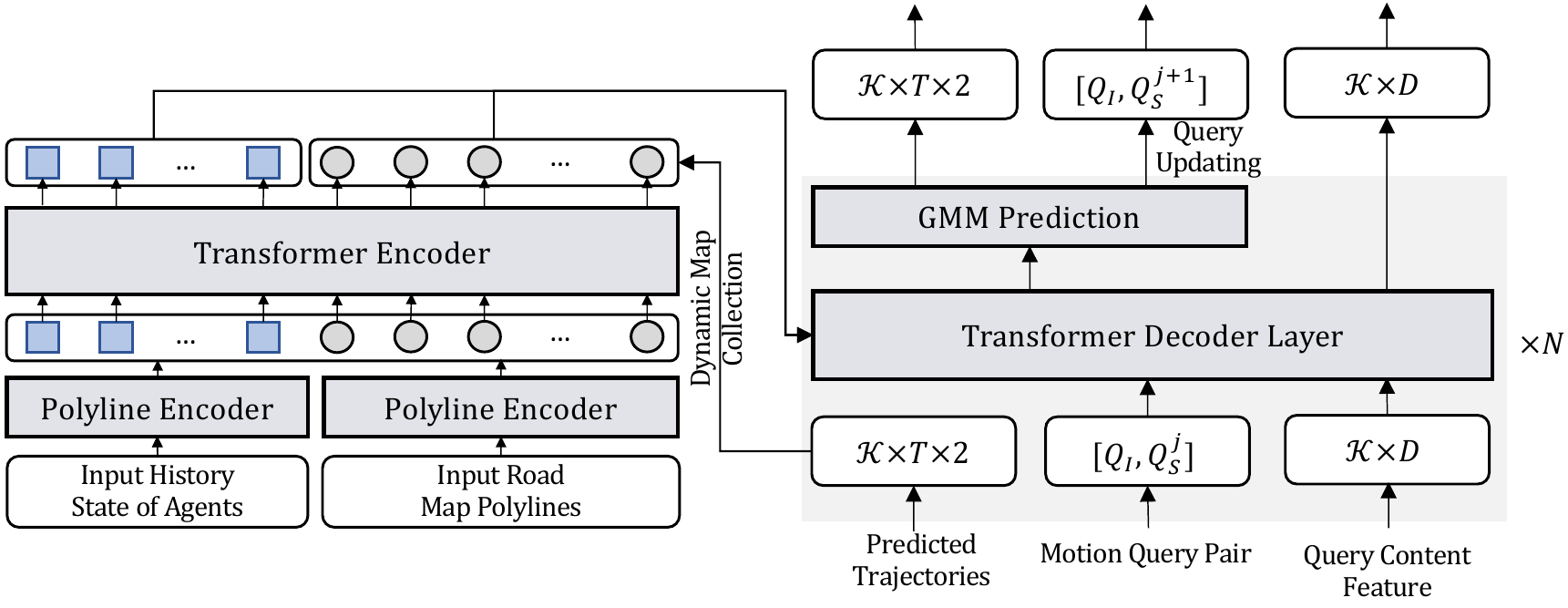}\\
	\vspace{-0.2in}
	\caption{
		The architecture of our proposed Motion Transformer framework for multimodal motion prediction.
	}
	\label{fig:framework}
	\vspace{-5mm}
\end{figure*}

\section{Method} \label{sec:method}\vspace{-1mm}
The overall architecture of our approach is shown in Fig.~\ref{fig:framework}, and it consists of a transformer encoder network for scene context encoding and a transformer decoder network for multimodal motion prediction.

\subsection{Context Encoding with Transformer Encoders}\vspace{-1mm}
To predict the future behavior of the agent, the first step is to model the interaction of all agents and encode the road environment. 
For this purpose, we adopt a simple and effective encoder network with stacked transformer encoders.

\myparagraph{Input representation.}
We adopt the agent-centric strategy as in \cite{zhao2020tnt,gu2021densetnt,varadarajan2021multipath++}, where both the agent history trajectories and the road map are normalized to the coordinate system centered at our interested agent. 
We utilize the vector representation~\cite{gao2020vectornet} to organize both agent's history trajectories and the road map as polylines.
A simple PointNet-like~\cite{qi2017pointnet} polyline encoder is adopted to encode these two input polyline representations, which produces the agent features $A \in \mathbb{R}^{N_a \times D}$ and the map features $M \in \mathbb{R}^{N_m \times D}$ ($N_a$ is the number of agent, $N_m$ is the number of map polylines and $D$ is the feature dimension).  

\myparagraph{Scene context encoding with transformer encoder.}
Given the encoded polyline features of the agents and the road maps, we adopt a simple encoder network with stacked transformer encoder layers to model the agent interaction and encode the road environment. It takes the agent features $A$ and map features $M$ as input, and a set of self-attention modules are then adopted on $A$ and $M$ to model the interaction of agent and also encode the scene environment features for the following decoder network.

\subsection{Multimodal Trajectory Prediction}\vspace{-1mm}
Given the encoded scene context features $A$ and $M$, a novel transformer-based decoder network is adopted for predicting multimodal future trajectories. 
Inspired by the concept of object query~\cite{carion2020end} for object detection, we propose the motion query pair to model motion prediction, which consists of a static intention query and a dynamic searching query, aiming at global intention localization and local movement refinement, respectively. 

\myparagraph{Motion query pair for motion prediction.}
Our motion query pair aims to localize the potential motion intentions of the agents and generate their future trajectories in a mode-specific manner. 
Hence, each motion query pair actually is associated with a specific intention point. 
Specifically, for each category, we generate $\mathcal{K}$ representation intention points  (denotaed as $I$) by using the k-means algorithm on the endpoints of ground-truth trajectories of the training set. 

Each static intention query is the learnable position embedding of a specific intention point by using a simple multi-layer perceptron (MLP) network, where each static intention query takes charge of predicting future trajectory for a specific motion mode. 
The static intention query is formulated as $Q_I=\text{MLP}(\text{PE}(I))$, where each static intention query is associated with a fixed intention point in different decoder layers, and $\text{PE}(\cdot)$ indicates the sinusoidal position encoding. 
This mode-specific motion prediction greatly stabilizes the training process and also facilitates predicting multimodal trajectories by enabling each motion mode to have their own learnable embedding. 

To complement with the static intention query for predicting better future trajectories, we further adopt a dynamic searching query for each static intention query, which aims at iteratively refining the predicted trajectory with updated fine-grained trajectory feature.
Specifically, the dynamic searching query is also initialized as a learnable position embedding of its corresponding intention point, but it will be dynamically updated based on the predicted trajectory in each decoder layer, so as to collect trajectory-specific features for iterative motion refinement. 
Hence, the dynamic searching query of $(j+1)$-th layer can be updated as $Q_S^{j+1} = \text{MLP}(\text{PE}(Y^j_T))$, where $Y^j_T$ is the endpoint of the predicted trajectory in $j$-th decoder layer, and $T$ is the number of future frames for trajectory prediction.
Moveover, we propose a dynamic map collection strategy to extract trajectory-aligned feature based on the predicted trajectory in $j$-th layer, which is implemented by collecting the closest 128 map polylines along the predicted trajectory (\emph{i.e.}, the map polylines are firstly ranked by calculating the smallest distance of its polyline center and all the predicted 80 waypoints of a single trajectory, and then we select the closest 128 map polylines for this trajectory). The collected fine-grained map features are then inputted to the $(j+1)$-th decoder layer for refining the predicted trajectory from $j$-th deocder layer.
Note that for the first decoder layer, the dynamic map collection for each dynamic searching query is implemented by collecting map polylines around its intention point.

\begin{table*}[t]
		\centering\small
		\setlength\tabcolsep{9pt}
		\caption{Top 10 entries on the test leaderboard of motion prediction track of 2022 Waymo Open Dataset Challenge.  Our approach is termed as MTR-A, \emph{i.e.}, Motion Transformer Advanced. The Soft mAP is the official ranking metric while the miss rate is the secondary ranking metric.}
			\begin{tabular}{l|ccccc}
				\hline
				Method & \cellcolor{mygray}{\bf Soft mAP}$\uparrow$ &  { mAP}~$\uparrow$ & minADE~$\downarrow$ & minFDE~$\downarrow$ & Miss Rate~$\downarrow$ \\ 
				\hline
				 MTR-A $1^{\text{st}}$ (Ours) & \cellcolor{mygray}{\bf 0.4594} & {\bf 0.4492} & 0.5640 & 1.1344 & {\bf 0.1160} \\ 
				 golfer $2^{\text{nd}}$ & \cellcolor{mygray}0.4259 & 0.4119 & {\bf 0.5533} & 1.1608 & 0.1354 \\ 
				 HBEns $3^{\text{rd}}$ & \cellcolor{mygray}0.3797 & 0.3700 & 0.6431 & 1.3405 & 0.1592 \\ 
				 (Null) & \cellcolor{mygray}0.3777 & 0.3719 & 0.6132 & 1.3218 & 0.1730 \\
				 DM & \cellcolor{mygray}0.3766 & 0.3710 & 0.6777 & 1.3558 & 0.1646 \\ 
				 HDGT(softmAP) & \cellcolor{mygray}0.3709 & 0.3577 & 0.7676 & {\bf 1.1077} & 0.1325 \\ 
				 MAML & \cellcolor{mygray}0.3445 & 0.3383 & 0.6945 & 1.4652 & 0.1846 \\ 
                 Gnet & \cellcolor{mygray}0.3367 & 0.3213 & 0.6255 & 1.2432 & 0.1740 \\ 
                 prenet & \cellcolor{mygray}0.3319 & 0.3168 & 0.6063 & 1.2415 & 0.1678 \\ 
                 HDGT & \cellcolor{mygray}0.3246 & 0.2826 & 0.5703 & 1.1434 & 0.1440 \\ 
				\hline
		\end{tabular}
		\label{tab:wod_test}
	\end{table*}
	
	  \begin{table*}[t]
		\centering\small
		\setlength\tabcolsep{9pt}
		\caption{Per-class performance of our approach on the validation set of Waymo Open Motion Dataset. }
			\begin{tabular}{c|c||cccc}
				\hline
				Setting & Category & { mAP}~$\uparrow$ & minADE~$\downarrow$ & minFDE~$\downarrow$ & Miss Rate~$\downarrow$ \\
				\hline
				\multirow{4}{*}{MTR} & Vehicle & 0.4620 &0.7559 & 	1.5229 & 0.1541 \\
				& Pedestrian & 0.429 & 0.3341 & 0.6881 & 0.0706	 \\
				& Cyclist &	0.3647 & 0.7037 & 1.4119 & 0.1802 \\
				& {\bf Avg} & 0.4186 & 0.5979 & 1.2076 & 0.1350\\
				\hline 
				\multirow{4}{*}{\tabincell{c}{MTR\\ (Ensemble)}} & Vehicle & 0.4911 & 0.6676 & 1.3331 & 0.1200\\
				& Pedestrian & 0.4550 &  0.3397 &	0.7077 & 0.0674	\\
				& Cyclist & 0.4191 & 0.6718 & 1.3489 & 0.1627	 \\
				& {\bf Avg} & 0.4551 & 0.5597 & 1.1299 & 0.1167\\
				\hline
		\end{tabular}
		\vspace{-3mm}
		\label{tab:wod_val}
	\end{table*}

\myparagraph{Attention with motion query pair.}
The transformer decoder takes each motion query pair as the query embedding, and aims to aggregate context information from both the agent features $A$ and the map features $M$. 
For each transformer decoder layer, we first utilize the static intention query to propagate information among different motion intentions by adopting the self-attention module, which generates the query content features for the following cross attention module.  
Then, the dynamic searching query is considered as the query embedding for cross attention module, and two separate cross attention modules are adopted for aggregating information from $A$ and $M$, respectively.  
These two aggregated features are concatenated as the queried features for each motion query pair, aiming at predicting future trajectory for its corresponding motion mode. 

\myparagraph{Motion prediction head with GMM.}
Given the queried feature for each motion query pair in each layer, we attach a simple prediction head with several MLP layers for predicting the future trajectory according to each queried features.
We follow \cite{chai2019multipath,varadarajan2021multipath++} to model the multimodal future motion with Gaussian Mixture Model (GMM) at each time step, where we predict a probability $p$ and GMM parameter $\mathcal{N}(\mu_x, \sigma_x; \mu_y, \sigma_y; \rho)$ for each motion query pair at each future time step.

The whole framework is optimized by adopting the negative log-likelihood loss to maximum the likelihood of ground-truth trajectory in each decoder layer. 
Inspired by the hard-assignment strategy~\cite{chai2019multipath,varadarajan2021multipath++}, we also select a positive Gaussian component from the predicted GMMs of $\mathcal{K}$ motion query pairs, where the selection is based on calculating the distance between each intention point and the endpoint of GT trajectory. 
The loss at each time step can be formulated as:
\begin{align}
    L_G&=-\log\mathcal{N}(\hat{Y}_x - \mu_x, \sigma_x; \hat{Y}_y - \mu_y, \sigma_y; \rho)
\end{align}
where $(\hat{Y}_x, \hat{Y}_y)$ is a waypoint of the selected ground-truth trajectory at this time step. 
The final loss is calculated by equally summing the loss of each decoder layer.  

\subsection{Model Ensemble}\label{sec:model_ensemble}\vspace{-1mm}
In order to further boost the performance of our framework, we adopt a model ensemble strategy to combine the results from multiple variants of our framework. 
Specifically, given $N_e$ well-trained models, we first collect $6$ predicted future trajectories from each model, which results in $6N_e$ multimodal future trajectories for each of our interested agent. Each trajectory has their own predicted confidence from their original model. 
We then select top $6$ future trajectories by adopting non-maximum-suppression (NMS) on the endpoints of these predicted trajectories, where the distance threshold $\delta$ is scaled along with the length $L$ of the trajectory that has the highest confidence among $6N_e$ predictions, as follows:
\begin{align}
\tiny
    \delta= \min\left(3.5,~ \max \left(2.5,~ \frac{L - 10}{50 - 10}\times 1.5 + 2.5\right)\right).
\end{align}
This simple model ensemble strategy facilitates taking the best predictions from multiple models, leading to better prediction of multimodal future trajectories. 
Note that our proposed MTR with this model ensemble strategy is denoted as MTR-A in the following experiment section.

\section{Experiments} \label{sec:intro}
\subsection{Implementation Details}

\myparagraph{Architecture details.} 
In the default setting of our model, we adopt 6 transformer encoder layers for the context encoding and 6 transformer decoder layers for generating the multimodal future trajectories. 
The hidden feature dimension is set to 512 to get a large model capacity for such a large-scale Waymo Open Motion Dataset (WOMD)~\cite{ettinger2021large}.   
For the context encoding, the road map is represented as polylines, where each polyline contains up to 20 map points (about $10m$ in WOMD). 
For the prediction head, a three-layer MLP head is adopted with feature dimention 512. We do not use any traffic light data in our model. 

For each category, we adopt 64 motion query pairs based on 64 intention points that are generated by k-means clustering algorithm on the training set. 
During testing, we adopt NMS with distance threshold $2.5m$ to select top 6 predictions from 64 predicted trajectories. 

\myparagraph{Training details.}
Our model is trained end-to-end by AdamW optimizer with a learning rate of 0.0001 and batch size of 80 scenes. 
All models are trained with 60 epochs, and we decay the learning rate by a factor of 0.5 every 5 epochs from epoch 30. The weight decay is set as 0.01 and we do not use any data augmentation. 
We utilize a single model to generate future trajectories for all three categories.

\myparagraph{Model ensemble details.}
We trained 7 variants of our model for conducting the model ensemble, where the variables includes the number of decoder layers (\emph{e.g.}, 6, 9), the number of motion queries (\emph{e.g.}, 6, 64, 100), and the hidden feature dimension (\emph{e.g.}, 256, 512). 
As mentioned in Sec.~\ref{sec:model_ensemble}, the predicted results of 7 models are finally combined with NMS to 
generate the final results.

\subsection{Main Results}
Table~\ref{tab:wod_test} shows the top 10 entries of the final leaderboard of 2022 Waymo Open Dataset Motion Prediction challenge. 
Our approach ranked 1st place on the leaderbaord, and surpasses all other submissions with remarkable margins in terms of Soft mAP, mAP and the miss rate, which demonstrates that our approach can predict better multimodal future trajectories. 
Besides that, as shown in Table~\ref{tab:wod_val}, we also report the per-class performance of our single-model results and the model ensemble results for reference.

\section{Conclusion}
In this technical report, we present a Motion Transformer framework for multimodal motion prediction. We adopt a small set of novel motion query pairs, where each motion query pair takes charge of predicting future trajectory of a specific motion mode to stabilize the training process and also generate better multimodal motion predictions.
Our proposed Motion Transformer achieves state-of-the-art performance for generating multimodal future trajectories, ranking 1st place  on the 2022 Waymo Open Dataset Motion Prediction Challenge.

{\small
\bibliographystyle{ieee_fullname}
\bibliography{egbib}

\begin{thebibliography}{10}\itemsep=-1pt

\bibitem{carion2020end}
Nicolas Carion, Francisco Massa, Gabriel Synnaeve, Nicolas Usunier, Alexander
  Kirillov, and Sergey Zagoruyko.
\newblock End-to-end object detection with transformers.
\newblock In {\em ECCV}, 2020.

\bibitem{chai2019multipath}
Yuning Chai, Benjamin Sapp, Mayank Bansal, and Dragomir Anguelov.
\newblock Multipath: Multiple probabilistic anchor trajectory hypotheses for
  behavior prediction.
\newblock In {\em CoRL}, 2019.

\bibitem{ettinger2021large}
Scott Ettinger, Shuyang Cheng, Benjamin Caine, Chenxi Liu, Hang Zhao, Sabeek
  Pradhan, Yuning Chai, Ben Sapp, Charles~R Qi, Yin Zhou, et~al.
\newblock Large scale interactive motion forecasting for autonomous driving:
  The waymo open motion dataset.
\newblock In {\em ICCV}, 2021.

\bibitem{gao2020vectornet}
Jiyang Gao, Chen Sun, Hang Zhao, Yi Shen, Dragomir Anguelov, Congcong Li, and
  Cordelia Schmid.
\newblock Vectornet: Encoding hd maps and agent dynamics from vectorized
  representation.
\newblock In {\em CVPR}, 2020.

\bibitem{gu2021densetnt}
Junru Gu, Chen Sun, and Hang Zhao.
\newblock Densetnt: End-to-end trajectory prediction from dense goal sets.
\newblock In {\em ICCV}, 2021.

\bibitem{jia2022hdgt}
Xiaosong Jia, Penghao Wu, Li Chen, Hongyang Li, Yu Liu, and Junchi Yan.
\newblock Hdgt: Heterogeneous driving graph transformer for multi-agent
  trajectory prediction via scene encoding.
\newblock In {\em arXiv preprint arXiv:2205.09753}, 2022.

\bibitem{liang2020learning}
Ming Liang, Bin Yang, Rui Hu, Yun Chen, Renjie Liao, Song Feng, and Raquel
  Urtasun.
\newblock Learning lane graph representations for motion forecasting.
\newblock In {\em ECCV}, 2020.

\bibitem{liu2021multimodal}
Yicheng Liu, Jinghuai Zhang, Liangji Fang, Qinhong Jiang, and Bolei Zhou.
\newblock Multimodal motion prediction with stacked transformers.
\newblock In {\em CVPR}, 2021.

\bibitem{ngiam2021scene}
Jiquan Ngiam, Vijay Vasudevan, Benjamin Caine, Zhengdong Zhang, Hao-Tien~Lewis
  Chiang, Jeffrey Ling, Rebecca Roelofs, Alex Bewley, Chenxi Liu, Ashish
  Venugopal, et~al.
\newblock Scene transformer: A unified architecture for predicting future
  trajectories of multiple agents.
\newblock In {\em ICLR}, 2022.

\bibitem{qi2017pointnet}
Charles~R Qi, Hao Su, Kaichun Mo, and Leonidas~J Guibas.
\newblock Pointnet: Deep learning on point sets for 3d classification and
  segmentation.
\newblock In {\em CVPR}, 2017.

\bibitem{tolstaya2021identifying}
Ekaterina Tolstaya, Reza Mahjourian, Carlton Downey, Balakrishnan Vadarajan,
  Benjamin Sapp, and Dragomir Anguelov.
\newblock Identifying driver interactions via conditional behavior prediction.
\newblock In {\em ICRA}, 2021.

\bibitem{varadarajan2021multipath++}
Balakrishnan Varadarajan, Ahmed Hefny, Avikalp Srivastava, Khaled~S Refaat,
  Nigamaa Nayakanti, Andre Cornman, Kan Chen, Bertrand Douillard, Chi~Pang Lam,
  Dragomir Anguelov, et~al.
\newblock Multipath++: Efficient information fusion and trajectory aggregation
  for behavior prediction.
\newblock In {\em ICRA}, 2022.

\bibitem{zhao2020tnt}
Hang Zhao, Jiyang Gao, Tian Lan, Chen Sun, Benjamin Sapp, Balakrishnan
  Varadarajan, Yue Shen, Yi Shen, Yuning Chai, Cordelia Schmid, et~al.
\newblock Tnt: Target-driven trajectory prediction.
\newblock In {\em CoRL}, 2020.

\end{thebibliography}
}

\end{document}